\documentclass[11pt,a4paper]{article}
\usepackage{amsfonts}
\usepackage{graphicx,subfigure} 
\usepackage{color}  
\usepackage{booktabs}

\usepackage{geometry}
\usepackage{coling2020}
\usepackage{times}
\usepackage{url}
\usepackage{latexsym}
\usepackage{microtype}

\usepackage{enumitem}
\setenumerate[1]{itemsep=0pt,partopsep=0pt,parsep=\parskip,topsep=5pt}
\setitemize[1]{itemsep=0pt,partopsep=0pt,parsep=\parskip,topsep=5pt}
\setdescription{itemsep=0pt,partopsep=0pt,parsep=\parskip,topsep=5pt}

\colingfinalcopy 

\def\modelName{ECNU-SenseMaker} 

\title{{\modelName} at SemEval-2020 Task 4: Leveraging  Heterogeneous Knowledge Resources for Commonsense Validation and Explanation}
\author{Qian Zhao \and Siyu Tao \and Jie Zhou \and Linlin Wang  \footnotemark[1] \and Xin Lin \footnotemark[1] \and Liang He\\
        School of Computer Science and Technology,\\
        East China Normal University, Shanghai 200062, China\\
        \{im0qianqian, sytao, jzhou\}@ica.stc.sh.cn\\
        \{llwang, xlin, lhe\}@cs.ecnu.edu.cn
        }
\date{}
\begin{document}
\maketitle

\renewcommand{\thefootnote}{\fnsymbol{footnote}}
\footnotetext[1]{Equal corresponding author. Email: xlin@cs.ecnu.edu.cn}
\blfootnote{This work is licensed under a Creative Commons Attribution 4.0 International License. License details: http://creativecommons.org/licenses/by/4.0/.}
\begin{abstract}
  This paper describes our system for SemEval-2020 Task 4: Commonsense Validation and Explanation \cite{wang-etal-2020-semeval}. We propose a novel Knowledge-enhanced Graph Attention Network (KEGAT) architecture for this task, leveraging heterogeneous knowledge from both the structured knowledge base (i.e. ConceptNet) and unstructured text to better improve the ability of a machine in commonsense understanding. This model has a powerful commonsense inference capability via utilizing suitable commonsense incorporation methods and upgraded data augmentation techniques. Besides, an internal sharing mechanism is cooperated to prohibit our model from insufficient and excessive reasoning for commonsense. As a result, this model performs quite well in both validation and explanation. For instance, it achieves state-of-the-art accuracy in the subtask called Commonsense Explanation (Multi-Choice). We officially name the system as  ECNU-SenseMaker. Code is publicly available at https://github.com/ECNU-ICA/{\modelName}.
\end{abstract}
\section{Introduction}
Commonsense reasoning is the process of making decisions by combining facts and beliefs with some basic knowledge appeared in daily life, which is fundamental to many Natural Language Understanding (NLU) tasks \cite{DBLP:journals/cogsci/Johnson-Laird80}. However, most existing models are quite weak in terms of commonsense acquisition and understanding compared with humans since commonsense is often expressed implicitly and constantly evolving \cite{wang-etal-2019-make}.
Thus, the commonsense problem is considered to be an important bottleneck of modern NLU systems \cite{DBLP:journals/cacm/DavisM15}.

Many attempts have been made to  empower machines with human abilities in commonsense understanding, and  several benchmarks have emerged to verify the commonsense reasoning capability, such as SQUABU \cite{Davis_2016} and CommonsenseQA \cite{Talmor2019CommonsenseQAAQ,lin-etal-2019-kagnet}. However, these benchmarks estimate commonsense indirectly and do not include corresponding explanations.
Therefore, the Commonsense Validation and Explanation (ComVE) benchmark is proposed which directly asks the real reasons behind decision making \cite{wang-etal-2020-semeval}. It consists of three subtasks, and we propose an entire system ({\modelName}) to solve the first two subtasks, namely, Commonsense Validation and Commonsense Explanation (Multi-Choice). In fact, the two different subtasks correlate to each other. Subtask A (Validation) requires the model to select which one is invalid between two given statements. While Subtask B (Explanation) aims to  identify one real reason from two other confusing options to explain why the statement is invalid.



Compared with the previous benchmarks, these subtasks are non-trivial because machines does not have the ability of children and adults in accumulating commonsense knowledge in daily life.
Since most questions in such subtasks require knowledge beyond the mentioned facts in text,  even trained over  millions of sentences,  there still exist a great gap behind human performance in commonsense reasoning, not to mention explanation capability \cite{wang-etal-2019-make}. Besides, many previous methods on commonsense reasoning have some important drawbacks which remain to be unsolved. For example,
the ways of combining knowledge  into deep learning architectures  are far from satisfaction, and it is an issue to balance the tradeoff between noise and the amount of incorporated commonsense from  knowledge base such as ConceptNet \cite{speer2017conceptnet}. In addition, researchers still consider this commonsense knowledge base  to be incomplete   despite it  has been evolved for a long time. The large majority of commonsense is generally recognized  to be implicitly expressed in unstructured  text  or human interactions in everyday life. Furthermore, some  pilot experiments have shown inference remains a challenging problem in the above  subtasks \cite{wang-etal-2019-make}.



To overcome these difficulties, we propose a novel Knowledge-enhanced Graph Attention Network (KEGAT) architecture for both Commonsense Validation and Explanation. Different from the ordinary Graph Attention Networks \cite{velickovic2018graph}, this model could leverage heterogeneous knowledge resources to improve its commonsense reasoning ability. On one hand, commonsense knowledge in the  structured knowledge base (i.e. ConceptNet)  is  well incorporated and  utilized  by the system with an elegant structural design as knowledge-enhanced embedding module. On the other, an upgraded data augmentation technique is put forward to empower the model with commonsense understanding capability based on the unlimited commonsense knowledge learnt from a large amount of unstructured text. Besides, an internal sharing mechanism is cooperated to prohibit our model from insufficient and excessive reasoning for commonsense. In this way, we build a novel Graph Neural Network based architecture for both validation and explanation  tasks, making more accurate inference for commonsense understanding. We officially name it as the ECNU-SenseMaker system.
In summary, the main contributions of our system are as follows:
\begin{itemize}
  \item We propose the Knowledge-enhanced Graph Attention Network to solve Commonsense  Validation and Explanation,  leveraging heterogeneous knowledge resources  for better commonsense reasoning.
  \item Our system utilizes an elegant embedding module to well incorporate commonsense from  structured knowledge bases, and design novel approaches to alleviate the noise caused by external knowledge.
  \item Our system gains valuable commonsense knowledge   based on a large amount of unstructured text via a novel data augmentation technique,  which also  improves  the robustness of our model.
  \item Last but not the least, this system uses  an internal sharing mechanism to indirectly guide the reasoning process, which greatly avoids insufficient and excessive commonsense inference.
\end{itemize}

\section{Problem Definition}
Formally, each instance in the ComVE task  consists of 5 sentences $\{S_1, S_2, R_1, R_2, R_3\}$, where $S_1$ and $S_2$ are two statements with similar expressions but only one of them conforms to commonsense,
and $R_1, R_2, R_3$ stand for three optional  explanations.   Subtask A, called Commonsense Validation, requires the model to identify the against-common-sense sentence $\hat S^-$ from the above two statements  $S_1$ and $S_2$, where $\hat S^-$ $\in$ \{$S_1$ ,$S_2$\}. While Subtask B, called Commonsense Explanation (Multi-Choice), requires the model to pick the most reasonable explanation $\hat R$ from the three given options $R_1, R_2, R_3$ for the against-common-sense statement  $\hat S^-$, where $\hat R$ $\in$ \{$R_1$ ,$R_2$, $R_3$\}.
Subtask C, called Commonsense Explanation (Generation), is similar to Subtask B, but it requires the model to generate a reasonable explanation $\hat R$.

\section{Methodology}
\label{method}
\begin{figure}[ht]
  \centering
  \includegraphics[scale=0.39]{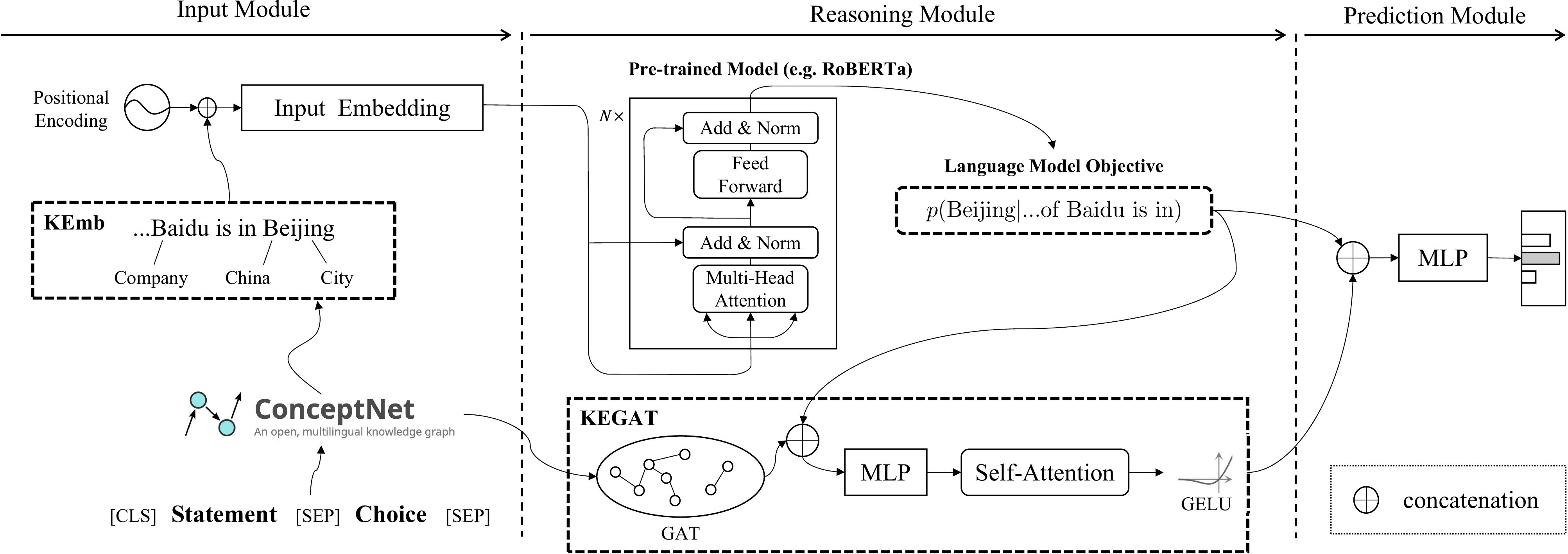}
  \caption{The overview of \modelName.}
  \label{fig:1}
\end{figure}

In this section, we describe the framework of our proposed {\modelName} system.  An overview of this architecture is depicted in Figure \ref{fig:1}.
We will introduce the basic components of this architecture, and provide some novel proposed strategies to enhance the
commonsense reasoning ability of our system.

\subsection{Input Module}
\begin{figure}[ht]
  \centering
  \includegraphics[scale=0.56]{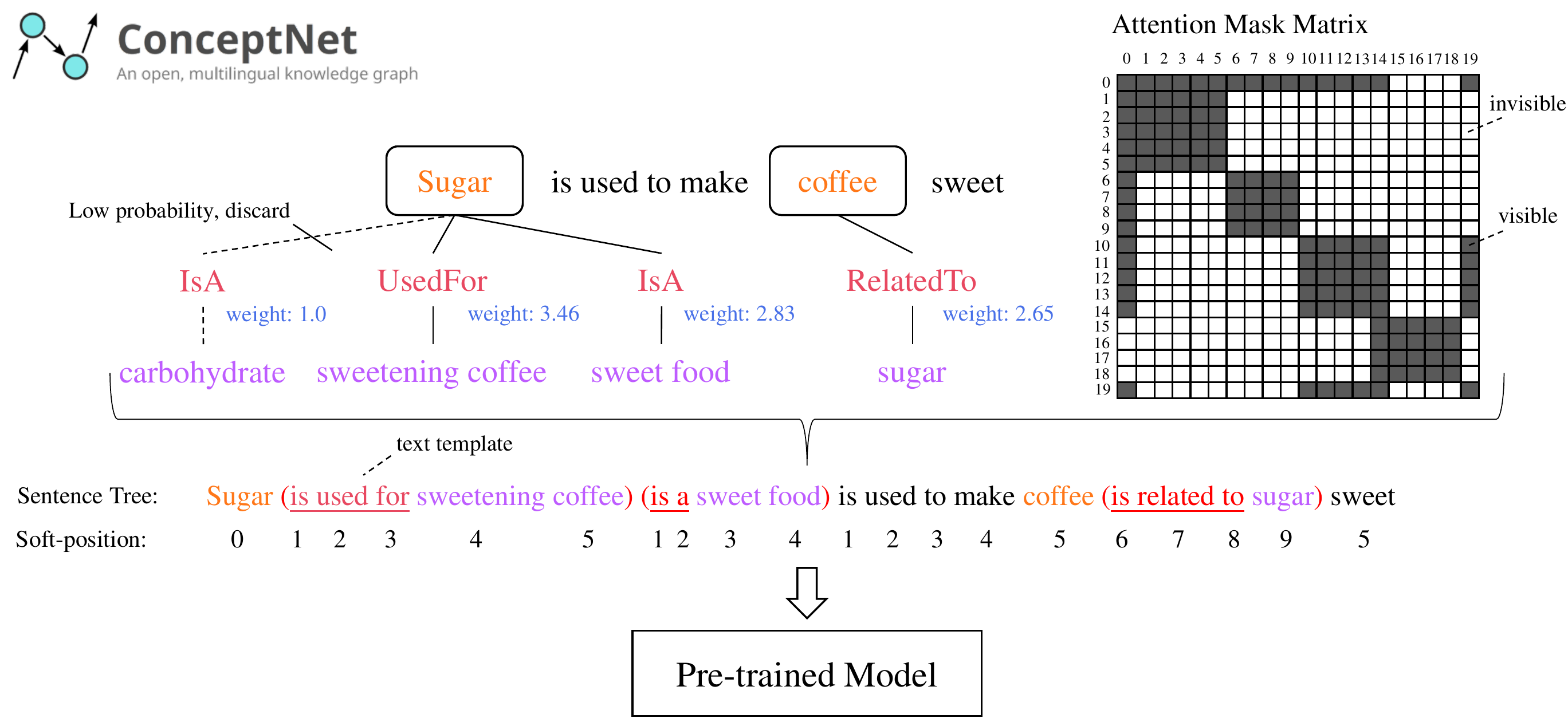}
  \caption{The KEmb module.}
  \label{fig:kbert}
\end{figure}

We cast both Subtask A and B as classification problems. To achieve this, every training instance in Subtask A is cast into  the form:
[CLS] $S_\alpha$ [SEP], where $\alpha \in\{1,2\}$. And we concatenate the against-common-sense statement $\hat S^-$ with every optional reason $R_\beta$
for Subtask B, which results in the form: [CLS] $\hat S^-$  [SEP] $R_\beta$ [SEP], where $\beta \in\{1,2,3\}$. For a specific training instance,
we utilize $U_i$ to denote the above converted form for convenience, where $i$ stands for the $i$-th statement or relation option given in each subtask.
Suppose that $A$ represents the option number of each subtask, we have $i \in \{1,2, \dots, A\}$, $A  \in \{|S|, |R|\}$,
and we assume that $y^*\in\{1, 2, \dots,A\}$ represents the label of the instance.
Then, we can adopt various methods to encode this converted input $U_i$. For instance, a basic way is to first
get the one-hot vector and positional encoding for all tokens, perform a  separate linearly transformation, and finally add these vectors to obtain an new embedding as $E_{U_i}$ for every $U_i$. Furthermore, we propose a more advanced approach to embed both the converted inputs and structured knowledge from external resources, which will be introduced in the next paragraph.
\paragraph{Structured Knowledge Incorporation}
\label{section:Structure Knowledge Incorporation}
Inspired by the work of K-BERT \cite{Liu2019KBERTEL}, we propose a novel Knowledge-enhanced Embedding (KEmb) module to well incorporate some structured knowledge from ConceptNet  \cite{speer2017conceptnet} to improve the commonsense understanding ability of the model. For all entities in every converted input $U_i$, we extract their adjacent entities  from  ConceptNet and add the corresponding relations to form a new tree-structured form, which is  depicted in the upper part of  Figure \ref{fig:kbert}.
We use soft-position instead of positional encoding, which is the relative distance between each node and the root node. We also use a attention mask matrix to maintain the invisibility of different branches in this tree-structured input and propose the following strategies to embed structured knowledge into the pre-trained model. Firstly, we use edge weight values in ConceptNet  as prior knowledge to select the most relevant adjacent entities and avoid some unnecessary noise since these weight values represent how believable the information is \cite{speer2017conceptnet}. Secondly, we manually design several text templates to describe the structured entities and their relation in ConceptNet with natural language, and insert the novel  pieces  of unstructured  text into the original input. An example is shown in Figure \ref{fig:kbert} to explain the detailed operations.
For  two entities ``sugar'' and ``coffee'' in this statement,  we  extract  their adjacent entities  from ConceptNet and select the most relevant ones based on the  corresponding edge weight values. Therefore, for the central entity ``sugar'', we only choose the high-correlated entities ``sweetening coffee'' and ``sweet food'',  but discard ``carbohydrate''  with a  low weight value. For the triple (sugar, UsedFor, sweetening coffee), we use our template to convert it to be ``Sugar is used to sweetening coffee" and insert this piece into the original statement with the soft-position and attention mask matrix operations \cite{Liu2019KBERTEL}. Finally, the flatten input can be handled by the pre-trained model described next.


\subsection{Reasoning Module}
To improve the reasoning ability of the model,  we propose to model the input  sentences  in a multi-level perspective which combines entities with the entire  statements or possible explanations.  First of all, we utilize the pre-trained model to process the above embedding $E_{U_i}$ obtained in the previous step to provide a high-level representation as
$\hat{E}_{U_i}^{base}$. This is not only because the pre-trained architectures
have achieved state-of-the-art performance in a variety of NLP tasks \cite{devlin-etal-2019-bert,NIPS2019_8812,Lan2019ALBERTAL}, but also because of the objective of language model, which can serve as a metric to estimate the possibility of a sentence being commonsensical. In this part, we choose an context-sensitive pre-trained model, RoBERTa \cite{Liu2019RoBERTaAR}, which is composed of $N$-layer transformer encoder \cite{NIPS2017_7181} depicted in the middle of Figure \ref{fig:1}. Secondly, we propose a Knowledge-enhanced Graph Attention Network (KEGAT) component to reason based on all relevant entities and the high-level representation of the entire statements or explanations from the pre-trained model. The principle of our proposed KEGAT is introduced next.

\paragraph{Knowledge-enhanced Graph Attention Network (KEGAT)}
As shown in Figure \ref{fig:1}, our KEGAT mainly consists of  a Graph Attention Network, a self-attention sub-module and several multi-layer perceptrons (MLP), which enables a multi-level reasoning from entities to sentences.

For the entity level, we still utilize some structured knowledge from ConceptNet but take a different incorporation approach,  which aims to
conduct commonsense  inferences over newly constructed  subgraphs. To achieve this, we adopt the N-gram method extract all entities  from the converted input
$U_i$, and use edge weight as the probability to sample a maximum of $k$ adjacent nodes from ConceptNet to form a subgraph for every extracted entity.  Suppose
the number of entities is  $n$, we have $n$ constructed subgraphs in total, which may be connected with edges. After that, we use  the conceptnet-numberbatch\footnote{ConceptNet-Numberbatch: https://github.com/commonsense/conceptnet-numberbatch} to get the $i$-th entity embeddings as the initial  representation $h_i^{(0)}$, which will be refined by the $L$-layer Graph Attention Network (GAT) \cite{velickovic2018graph}. In the refinement process,  the GAT module automatically learns a optimal edge weight  between two entities in these subgraphs based on the Commonsense Validation or Explanation subtasks, indicating the relevance of adjacent entities to every central entity. In other words, for a central entity,  the GAT attempts to  only assign higher weight values to those edges connected  with several most commonsensical adjacent entities from the constructed subgraph, and discard some irreverent edges. Thus, the commonsense inference ability of our model is highly improved with those knowledge incorporated by the refined subgraphs.  The working principles of our GAT is in Eq. \ref{eq:1}--\ref{eq:2}.
\begin{equation}  
  \label{eq:1}
  h_i^{(l+1)}=\sigma\Bigg(\frac{1}{M}\sum_{m=1}^{M}\sum_{j\in\mathcal{N}_i}\alpha_{{ij}_m}^{(l)}\textbf{W}^{(l)}_{m}h_{j}^{(l)}\Bigg)
\end{equation}
\begin{equation}
  \label{eq:1r1}
  \alpha_{ij}^{(l)}={\rm softmax}_j\Big({\rm f}\big([\textbf{W}^{(l)}h_i^{(l)}; \textbf{W}^{(l)}h_j^{(l)}]\big)\Big)
\end{equation}
We update every entity node  based on  Eq. \ref{eq:1}, where
$\sigma(\cdot)$ stands for a ELU function \cite{DBLP:journals/corr/ClevertUH15},
$\textbf{W}$ is the network parameter, $h_i^{(l)}$ is the representation from the $l$-th layer  of GAT, and
$\mathcal{N}_i$ represents all adjacent nodes to the $i$-th entity.
$M$ is the number of independent attention mechanisms in Eq. \ref{eq:1r1}, and
$a_{ij}^{(l)}$ represents  the relevance degree of the $j$-th adjacent entity respect to the $i$-th  entity.
Besides,  ${\rm f(\cdot)}$ is a projection function converting vector to real number, and $[;]$ stands for the  concatenation operation. Finally , we define
\begin{equation}
  \label{eq:2}
  \hat{E}_{U_i}^{gnn}=\frac{1}{n}\sum_{i=1}^{n}{h_i^{(L)}}
\end{equation}
to be the final representation for entity subgraphs obtained from the GAT.

From the sentence level, we propose to adopt a self-attention submodule and many MLPs to promote the model to reason over both entities and input sentences.
We first utilize a MLP to fuse the symbolic and semantic representations and then take a self-attention operation for refinement. Thus, the entity-level representation can be
further refined taking the statements and possible explanations as a reference. In a word, some valuable dimensions  can be highlighted to retain the most commonsensical information
from the fused representations  $\hat{E}_{U_i}^{all}$ to the improve reasoning ability.
We formulate these operations in Eq. \ref{eq:3}--\ref{eq:4}.
\begin{equation}
  \label{eq:3}
  \hat{E}_{U_i}^{all} = {\rm MLP}([\hat{E}_{U_i}^{base}; \hat{E}_{U_i}^{gnn}])
\end{equation}
\begin{equation}
  \label{eq:4}
  G_{U_i} = \sigma({\rm SelfAttn}(\hat{E}_{U_i}^{all}))
\end{equation}
where $G_{U_i}$ is the refined representation,  SelfAttn($\cdot$)  stands for a self-attention operation,  and
$\sigma(\cdot)$ is the  activation function.  Finally, we concatenate  $G_{U_i}$ and $\hat{E}_{U_i}^{base}$ to obtain the entire reasoning representation.
\begin{equation}
  \label{eq:5}
  \hat{E}_{U_i} = [G_{U_i}; \hat{E}_{U_i}^{base}]
\end{equation}

\paragraph{Unstructured Knowledge Augmentation}
\label{section:Data Augmentation with via unstructured text}
Since commonsense knowledge may be scattered in a large amount of unstructured text, we introduce three novel types of unstructured resources to teach our {\modelName} system, aiming to improve the intelligence of the model in commonsense understanding. The first type comes from the training set of CommonsenseQA \cite{Talmor2019CommonsenseQAAQ}, and the second one originates from ConceptNet.  To be detailed, we manually design some templates to express the extracted relations from ConceptNet with natural language, in which way we generate a great many of unstructured text based on this structured commonsense knowledge base. As a matter of fact, these unstructured data share a similar distribution with the text in the ComVE dataset \cite{wang-etal-2020-semeval}. Therefore, we can use them to train our system in advance before the whole official process starts, which always leads to a better initial weight  for our model. Apart from this, we integrate some data from Subtask C to serve as the third type of unstructured resources to increase the generalization ability of our system, which is  allowed by competition regulations.

\subsection{Prediction Module}
\label{section:Prediction Module}
After the above multi-level reasoning, for every training data  we obtain the representation $\{\hat{E}_{U_i}\}_{i=1}^{A}$ of converted inputs. In the prediction module, we use a multilayer perceptron (MLP) to solve the downstream tasks of commonsense validation and explanation based on Eq. \ref{eq:p0}-- \ref{eq:p1}.
\begin{equation}
  \label{eq:p0}
  P_i = {\rm MLP}(\hat{E}_{U_i}),~~
  P' = {\rm softmax}(P), ~~
  y = \arg\max(P')
\end{equation}
\begin{equation}
  \label{eq:p1}
  \mathcal{L} = -\sum_{i=1}^{A}{y^*_i\log P'_i}
\end{equation}
where $P$ is the output of the MLP, and  $P \in \mathbb{R}^{A \times 1}$.  $y$ stands for the prediction result, $P'_i$ represents the probability of selecting the $i$-th statement label or
explanation. The training objective is in Eq  \ref{eq:p1}, which aims to minimize negative log-likelihood, and $y^*$ here stands for one-hot vector of the optimal label.


\subsection{Adaptive  Strategies}
\paragraph{Noise Alleviation Strategy}
Previous knowledge incorporation approaches often lead to inevitable noise \cite{10.1007/978-3-030-32233-5_2,DBLP:conf/aaai/WangKMYTACFMMW19}, and it is still  an open research question  to balance the tradeoff between noise and the amount of incorporated commonsense from  knowledge base consisting of all types of entity nodes \cite{Weissenborn2018DynamicIO,DBLP:conf/conll/KhashabiKSR17}. The proposed KEmb and KEGAT have provided possible solutions to  alleviate the noise caused by incorporated
structured knowledge to some extent. These two modules share a same goal of identifying the most commonsensical external entities and discarding the irreverent ones. Take the KEGAT for instance, we achieve this based on both entity-level and sentence-level inference  thoroughly described in the previous  Reasoning Module part. Furthermore, several unimportant types of edges have been removed to avoid unnecessary noises, such as ``/r/ExternalURL'', ``/r/DistinctFrom'', etc.

\paragraph{Internal Sharing Mechanism}
\label{section:Internal Sharing Mechanism}
To further improve the commonsense inference ability  of our system, we propose an internal sharing mechanism which utilizes some processed outputs of  pre-trained models to  assist  the KEGAT module in reasoning  with a  right direction, aiming to avoid insufficient or excessive inferences. First of all, we add a new multilayer perceptron  to process the output of the RoBERTa pooler layer $\hat{E}_{U_i}^{base}$,  the representation from  pre-trained models.  Meanwhile, we can also get  the embedding of every input token, and  project this embedding to be a vocabulary-size vector with the network. In this way, it is easy to utilize the idea of  classification problems to calculate the loss for every token,  which measures the difference between the converted embedding to the original input.  Thus, we sum up all these to get the overall loss of the sentence, and utilize CrossEntropy to obtain  the pre-trained module loss  as $\mathcal{L}_1$.  We use a similar approach to \newcite{DBLP:journals/corr/KendallGC17} to combine $\mathcal{L}_1$ and the KEGAT loss $\mathcal{L}_2$. Therefore, we obtain a novel loss as
\begin{equation}
  \label{eq:6}
  \mathcal{L} = \frac{1}{2\sigma_{1}^{2}}\mathcal{L}_1(\textbf{W}) +
  \frac{1}{2\sigma_{2}^{2}}\mathcal{L}_2(\textbf{W}) + \log{\sigma_1\sigma_2}
\end{equation}
where $\sigma_1$ and $\sigma_2$ are two learnable parameters, they dynamically adjust the learning rate according to the difficulty of data fitting for better learning.
With the above internal sharing mechanism,  we can utilize the cross-entropy loss from the pre-trained model part  to guide the KEGAT module to decide whether to remove the relevant evidence chain or not.  Experiments show that this approach has a positive impact on the performance of our model in the commonsense validation and explanation task.

\section{Experiments}
\label{section:experiments}
\subsection{Datasets and Metric}
In the ComVE benckmark, Subtask A (Validation) requires the model to find the against-common-sense statement, while Subtask B (Explanation: Multi-Choice) aims to select the most reasonable explanation for the invalid statement. Examples of both subtasks are listed in Table \ref{tab:d0}. In both Subtask A and B,
we utilize Accuracy as a metric to evaluate model performance \cite{wang-etal-2020-semeval}.
\begin{table}[]
  \centering
  \begin{tabular}{@{}cl@{}}
    \toprule
    \textbf{Task} & \multicolumn{1}{c}{\textbf{Examples}} \\ \midrule
    Subtask  A    & \begin{tabular}[c]{@{}l@{}}\textbf{Which statement of the two is against commonsense?}\\S1: he put an elephant into the fridge. $\times$\\S2: he put a turkey into the fridge. $\bigcirc$\end{tabular}            \\ \midrule
    Subtask  B    & \begin{tabular}[c]{@{}l@{}}\textbf{Why is ``he put an elephant into the fridge'' against commonsense?} \\A. an elephant is much bigger than a fridge. $\surd$\\B. elephants are usually gray while fridges are usually white. $\times$\\C. an elephant cannot eat a fridge. $\times$\end{tabular}            \\ \bottomrule
  \end{tabular}
  \caption{Data description.}
  \label{tab:d0}
\end{table}
\subsection{Experimental Settings}
\label{section:experimental settings}
In our experiment, we set the batch size as 2 and maximum sentence length as 128.
During training, we freeze all layers except the last classification layer and learn 4 epochs with a learning rate of 0.001.
In the fine-tuning phase, we unfreeze all layers and learn 8 epochs with a learning rate of 0.000005.
Like the training phase, using the weights of the pre-trained language model itself to classify downstream tasks is beneficial to correct the randomly initialized classification layer.
Therefore, all layers of the entire model in the fine-tuning phase are suitable for fine-tuning with the same low learning rate.
In each phase, we save the model parameters at the time of the highest test accuracy and load it at the beginning of the next phase.
In addition, we use the Adam optimizer \cite{DBLP:journals/corr/KingmaB14} and set epsilon to 0.000001 for gradient descent.





\begin{table}[]
  \centering
  \begin{tabular}{@{}clcclc@{}}
    \toprule
    \multicolumn{3}{c|}{\textbf{Subtask A}} & \multicolumn{3}{c}{\textbf{Subtask B}}                                                                                                                       \\
    \textbf{Rank}                           & \multicolumn{1}{c}{\textbf{Team Name}} & \multicolumn{1}{c|}{\textbf{Accuracy}} & \textbf{Rank} & \multicolumn{1}{c}{\textbf{Team Name}} & \textbf{Accuracy} \\ \midrule
    1                                       & hit itnlp                              & \multicolumn{1}{c|}{97.00}             & \textbf{1}    & \textbf{ECNU ICA (Ours)}               & \textbf{95.00}    \\
    \textbf{2}                              & \textbf{ECNU ICA (Ours)}               & \multicolumn{1}{c|}{\textbf{96.70}}    & 2             & hit itnlp                              & 94.80             \\
    3                                       & iie-nlp-NUT                            & \multicolumn{1}{c|}{96.40}             & 3             & iie-nlp-NUT                            & 94.30             \\
    4                                       & nlpx                                   & \multicolumn{1}{c|}{96.40}             & 4             & Solomon                                & 94.00             \\
    5                                       & Solomon                                & \multicolumn{1}{c|}{96.00}             & 5             & NEUKG                                  & 93.80             \\ \midrule
    \multicolumn{6}{c}{\textbf{Baseline}}                                                                                                                                                                  \\ \midrule
    -                                       & BERT base                              & \multicolumn{1}{c|}{71.20}             & -             & BERT base                              & 62.10             \\
    -                                       & fine-tuned BERT base                   & \multicolumn{1}{c|}{89.10}             & -             & fine-tuned BERT base                   & 82.30             \\ \bottomrule
  \end{tabular}
  \caption{Top 5 results and baseline performance.}
  \label{tab:3}
\end{table}

\begin{table}[t]
  \centering
  \begin{tabular}{@{}lcc@{}}
    \toprule
    \textbf{Model}                                    & \textbf{Dev Acc.(\%)} & \textbf{Test Acc.(\%)} \\
    \midrule
    Random guess                                      & 50.00                 & 50.00                  \\
    \midrule
    \midrule
    fine-tuned BERT base                              & -                     & 89.10                  \\
    RoBERTa-large                                     & 94.68                 & 94.50                  \\
    + LM                                              & 96.19                 & 95.30                  \\
    + KEmb                                            & 96.39                 & 95.40                  \\
    + KEGAT                                           & 95.99                 & 95.40                  \\
    + LM + KEmb                                       & 96.39                 & 95.30                  \\
    + LM + KEGAT                                      & 95.89                 & 95.20                  \\
    + KEmb + KEGAT                                    & 96.19                 & 95.90                  \\
    + LM + KEmb + KEGAT                               & 96.09                 & 95.30                  \\
    + CommonsenseQA pre-trained                       & 95.49                 & 94.00                  \\
    \midrule
    \midrule
    RoBERTa-large + ALBERT-xxlarge                    & 95.90                 & 95.80                  \\
    RoBERTa-large LM + RoBERTa-large + ALBERT-xxlarge & \textbf{96.60}        & \textbf{96.70}         \\
    \midrule\midrule
    Human Performance \cite{wang-etal-2019-make}      & -                     & 99.10                  \\
    \bottomrule
  \end{tabular}
  \caption{Experimental results of ComVE Subtask A.
  }
  \label{tab:1}
\end{table}

\begin{table}[]
  \centering
  \begin{tabular}{@{}lcc@{}}
    \toprule
    \textbf{Model}                                    & \textbf{Dev Acc.(\%)} & \textbf{Test Acc.(\%)} \\
    \midrule
    Random guess                                      & 33.33                 & 33.33                  \\
    \midrule
    \midrule
    fine-tuned BERT base                              & -                     & 82.30                  \\
    RoBERTa-large                                     & 91.13                 & 92.90                  \\
    + LM                                              & 92.18                 & 93.70                  \\
    + KEmb                                            & 92.37                 & 91.90                  \\
    + KEGAT                                           & 92.78                 & 93.30                  \\
    + LM + KEmb                                       & 91.27                 & 91.90                  \\
    + LM + KEGAT                                      & 92.68                 & 93.00                  \\
    + KEmb + KEGAT                                    & 91.57                 & 92.30                  \\
    + LM + KEmb + KEGAT                               & 91.98                 & 91.80                  \\
    + CommonsenseQA pre-trained                       & 93.58                 & 93.60                  \\
    \midrule
    \midrule
    RoBERTa-large + ALBERT-xxlarge                    & 94.08                 & 94.00                  \\
    RoBERTa-large LM + ALBERT-xxlarge                 & 94.38                 & 94.60                  \\
    RoBERTa-large LM + RoBERTa-large + ALBERT-xxlarge & \textbf{94.68}        & \textbf{95.00}         \\
    \midrule\midrule
    Human Performance \cite{wang-etal-2019-make}      & -                     & 97.80                  \\
    \bottomrule
  \end{tabular}
  \caption{Experimental results of ComVE Subtask B.}
  \label{tab:2}
\end{table}
\subsection{Results}
Table \ref{tab:3} shows the results of the top five teams from the leadboard for Subtask A \& B (by  March 17). 
Our system achieves state-of-the-art accuracy in Subtask B. It outperforms the baseline (BERT base) model~\cite{wang-etal-2019-make} with a relative improvement of 52.98\% and achieves a  relative improvement of 15.43\% compared with fine-tuned BERT base. Meanwhile, experimental results also show that our system performs well in solving  Subtask A. 
Therefore, we conclude from Table~\ref{tab:3} that  our system has the ability of solving both commonsense validation and explanation tasks. 

Besides, we  test the performance of our system with the strategies mentioned in Section~\ref{method}. Here, ``+KEmb" stands for our system with the Knowledge-enhanced Embedding, ``+KEGAT" represents our proposed model with  Knowledge-enhanced Graph Attention Network, the  abbreviated name ``+LM "  refers to our system with the Internal Sharing Mechanism, and ``+CommonsenseQA pre-trained" stands for the system with data  augmentation technique. In addition, Dev Acc. and Test Acc. stand for the accuracy on the dev set and test set respectively.  Table \ref{tab:1} shows the experimental results of our system 
on Subtask A.  From this table, we conclude that on the test set, our system achieves  a relative improvement of 7.63\% when adding both the KEmb and KEGAT submodules compared with  the fine-tuned BERT base. Moreover, we also test the performance of two ensemble models shown  in the bottom of Table~\ref{tab:1}, and the ``RoBERTa-large LM + RoBERTa-large + ALBERT-xxlarge" ensemble obtains the best performance on the test set, which outperforms the fine-tuned BERT base model with a relative improvement of 8.53\%.  Here, this ensemble  is the combination of three models and the ``RoBERTa-large LM" stands for the RoBERTa-large model with the Internal Sharing Mechanism. Meanwhile, 
the results on Subtask B are shown in Table \ref{tab:2}, and we conclude from this table that our system with LM strategy outperforms the fine-tuned BERT base with a relative improvement of 13.85\%. Furthermore, our ensemble model achieves a relative improvement of 15.43\% compared with the fine-tuned BERT base.

Therefore, it can be concluded that  the ensemble models with the Internal Sharing Mechanism strategy gereatly improve the commonsense reasoning ability of our system, and the single system with multiple strategies performs well in most cases.  Besides, data augmentation techniques  proves to be an effective way to improve the performance of machine in commonsense validation and explanation. 
\subsection{Error Analysis}
To further improve the performance  in the future, it is helpful to study the failure cases of our model. In particular, we have categorized the observed
failure cases  in the  ComVE into the following categories.
\begin{enumerate}
  \item Decision or interpretation from different perspectives. For example, ``\emph{$S_1$: A war is fought for solution. $S_2$: There is peace during war.}'', where $S_2$ violates commonsense. And ``\emph{$S_1$: Humans wrote the bible about god. $S_2$: God lives physically on earth.}''  In this given example,  $S_2$ violates commonsense. Whereas there may exist various decision-making standards, under which both of the  two statements  should make sense from the point of view of humans.
  \item Ambiguous explanations are difficult to understand. For example, ``\emph{$\hat S^-$: My son had us write an essay on The National Monument. $R_1$: My son can write the alphabet instead of teach how to write an essay. $R_2$: My son isn't smart enough to assign an essay. $R_3$: My son cannot read and write an essay.}'' In this example, all three options, which seem to be less related and ambiguous, are expressed in a neutral way. Therefore, it would be more difficult for the machine to understand and make decisions.
\end{enumerate}
\section{Conclusion}
In this paper, we propose the {\modelName} system, which utilizes a novel Knowledge-enhanced Graph Attention Network architecture
to solve both commonsense validation and explanation tasks. It well incorporates heterogeneous knowledge from both unstructured text and structured resources such as ConceptNet, and relies on the
graph attention module with an internal sharing mechanism to improve the commonsense reasoning ability of the model.
Our model achieves good results in the newest leaderboard of SemEval-2020, solving two subtasks with an entire model.  We hope our work can shed some lights to  directly  empower machines with human abilities in commonsense understanding and reasoning.

\section*{Acknowledgements}
This work was supported by National Key R\&D Program of
China (No. 2018AAA0100503 \& No. 2018AAA0100500), National
Natural Science Foundation of China (No. 61773167), Science
and Technology Commission of Shanghai Municipality (No.19511120200), and Shanghai Sailing Program (No.20YF1411800).
\bibliographystyle{coling}
\bibliography{semeval2020}

\begin{thebibliography}{}

\bibitem[\protect\citename{Bosselut \bgroup et al.\egroup
  }2019]{bosselut-etal-2019-comet}
Antoine Bosselut, Hannah Rashkin, Maarten Sap, Chaitanya Malaviya, Asli
  Celikyilmaz, and Yejin Choi.
\newblock 2019.
\newblock {{COMET}: Commonsense Transformers for Automatic Knowledge Graph
  Construction}.
\newblock In {\em Proceedings of the 57th Annual Meeting of the Association for
  Computational Linguistics}, pages 4762--4779, Florence, Italy, July.
  Association for Computational Linguistics.

\bibitem[\protect\citename{Clevert \bgroup et al.\egroup
  }2016]{DBLP:journals/corr/ClevertUH15}
Djork{-}Arn{\'{e}} Clevert, Thomas Unterthiner, and Sepp Hochreiter.
\newblock 2016.
\newblock {Fast and Accurate Deep Network Learning by Exponential Linear Units
  (ELUs)}.
\newblock In Yoshua Bengio and Yann LeCun, editors, {\em 4th International
  Conference on Learning Representations, {ICLR} 2016, San Juan, Puerto Rico,
  May 2-4, 2016, Conference Track Proceedings}.

\bibitem[\protect\citename{Davis and Marcus}2015]{DBLP:journals/cacm/DavisM15}
Ernest Davis and Gary Marcus.
\newblock 2015.
\newblock {Commonsense Reasoning and Commonsense Knowledge in Artificial
  Intelligence}.
\newblock {\em Commun. {ACM}}, 58(9):92--103.

\bibitem[\protect\citename{Davis}2016]{Davis_2016}
Ernest Davis.
\newblock 2016.
\newblock {How to Write Science Questions that Are Easy for People and Hard for
  Computers}.
\newblock {\em AI Magazine}, 37(1):13--22, Apr.

\bibitem[\protect\citename{Devlin \bgroup et al.\egroup
  }2019]{devlin-etal-2019-bert}
Jacob Devlin, Ming-Wei Chang, Kenton Lee, and Kristina Toutanova.
\newblock 2019.
\newblock {{BERT}: Pre-training of Deep Bidirectional Transformers for Language
  Understanding}.
\newblock In {\em Proceedings of the 2019 Conference of the North {A}merican
  Chapter of the Association for Computational Linguistics: Human Language
  Technologies, Volume 1 (Long and Short Papers)}, pages 4171--4186,
  Minneapolis, Minnesota, June. Association for Computational Linguistics.

\bibitem[\protect\citename{Johnson{-}Laird}1980]{DBLP:journals/cogsci/Johnson-Laird80}
Philip~N. Johnson{-}Laird.
\newblock 1980.
\newblock {Mental Models in Cognitive Science}.
\newblock {\em Cognitive Science}, 4(1):71--115.

\bibitem[\protect\citename{Kendall \bgroup et al.\egroup
  }2018]{DBLP:journals/corr/KendallGC17}
Alex Kendall, Yarin Gal, and Roberto Cipolla.
\newblock 2018.
\newblock {Multi-Task Learning Using Uncertainty to Weigh Losses for Scene
  Geometry and Semantics}.
\newblock In {\em The IEEE Conference on Computer Vision and Pattern
  Recognition (CVPR)}, pages 7482--7491, 06.

\bibitem[\protect\citename{Khashabi \bgroup et al.\egroup
  }2017]{DBLP:conf/conll/KhashabiKSR17}
Daniel Khashabi, Tushar Khot, Ashish Sabharwal, and Dan Roth.
\newblock 2017.
\newblock {Learning What is Essential in Questions}.
\newblock In Roger Levy and Lucia Specia, editors, {\em Proceedings of the 21st
  Conference on Computational Natural Language Learning (CoNLL 2017),
  Vancouver, Canada, August 3-4, 2017}, pages 80--89. Association for
  Computational Linguistics.

\bibitem[\protect\citename{Kingma and Ba}2015]{DBLP:journals/corr/KingmaB14}
Diederik~P. Kingma and Jimmy Ba.
\newblock 2015.
\newblock {Adam: {A} Method for Stochastic Optimization}.
\newblock In Yoshua Bengio and Yann LeCun, editors, {\em 3rd International
  Conference on Learning Representations, {ICLR} 2015, San Diego, CA, USA, May
  7-9, 2015, Conference Track Proceedings}.

\bibitem[\protect\citename{Kipf and Welling}2017]{kipf2017semi}
Thomas~N. Kipf and Max Welling.
\newblock 2017.
\newblock {Semi-Supervised Classification with Graph Convolutional Networks}.
\newblock In {\em International Conference on Learning Representations (ICLR)}.

\bibitem[\protect\citename{Lan \bgroup et al.\egroup }2020]{Lan2019ALBERTAL}
Zhen-Zhong Lan, Mingda Chen, Sebastian Goodman, Kevin Gimpel, Piyush Sharma,
  and Radu Soricut.
\newblock 2020.
\newblock {ALBERT: A Lite BERT for Self-supervised Learning of Language
  Representations}.
\newblock {\em International Conference on Learning Representations (ICLR)},
  abs/1909.11942.

\bibitem[\protect\citename{Li \bgroup et al.\egroup
  }2019]{DBLP:journals/corr/abs-1909-09743}
Shiyang Li, Jianshu Chen, and Dian Yu.
\newblock 2019.
\newblock {Teaching Pretrained Models with Commonsense Reasoning: {A}
  Preliminary KB-Based Approach}.
\newblock In {\em Knowledge Representation \& Reasoning Meets Machine
  Learning}.

\bibitem[\protect\citename{Lin \bgroup et al.\egroup
  }2019]{lin-etal-2019-kagnet}
Bill~Yuchen Lin, Xinyue Chen, Jamin Chen, and Xiang Ren.
\newblock 2019.
\newblock {{K}ag{N}et: Knowledge-Aware Graph Networks for Commonsense
  Reasoning}.
\newblock In {\em Proceedings of the 2019 Conference on Empirical Methods in
  Natural Language Processing and the 9th International Joint Conference on
  Natural Language Processing (EMNLP-IJCNLP)}, pages 2829--2839, Hong Kong,
  China, November. Association for Computational Linguistics.

\bibitem[\protect\citename{Liu \bgroup et al.\egroup }2019]{Liu2019RoBERTaAR}
Yinhan Liu, Myle Ott, Naman Goyal, Jingfei Du, Mandar Joshi, Danqi Chen, Omer
  Levy, Mike Lewis, Luke Zettlemoyer, and Veselin Stoyanov.
\newblock 2019.
\newblock {RoBERTa: A Robustly Optimized BERT Pretraining Approach}.
\newblock {\em ArXiv}, abs/1907.11692.

\bibitem[\protect\citename{Liu \bgroup et al.\egroup }2020]{Liu2019KBERTEL}
Weijie Liu, Peng Zhou, Zhe Zhao, Zhiruo Wang, Qi~Ju, Haotang Deng, and Ping
  Wang.
\newblock 2020.
\newblock {{K-BERT}: Enabling Language Representation with Knowledge Graph}.
\newblock In {\em Proceedings of AAAI 2020}.

\bibitem[\protect\citename{{Mikolov} \bgroup et al.\egroup
  }2013]{2013arXiv1301.3781M}
Tomas {Mikolov}, Kai {Chen}, Greg {Corrado}, and Jeffrey {Dean}.
\newblock 2013.
\newblock {Efficient Estimation of Word Representations in Vector Space}.
\newblock {\em arXiv e-prints}, page arXiv:1301.3781, January.

\bibitem[\protect\citename{Speer \bgroup et al.\egroup
  }2017]{speer2017conceptnet}
Robyn Speer, Joshua Chin, and Catherine Havasi.
\newblock 2017.
\newblock {ConceptNet 5.5: An Open Multilingual Graph of General Knowledge}.
\newblock In {\em AAAI Conference on Artificial Intelligence}, pages
  4444--4451.

\bibitem[\protect\citename{Talmor \bgroup et al.\egroup
  }2019]{Talmor2019CommonsenseQAAQ}
Alon Talmor, Jonathan Herzig, Nicholas Lourie, and Jonathan Berant.
\newblock 2019.
\newblock {{C}ommonsense{QA}: A Question Answering Challenge Targeting
  Commonsense Knowledge}.
\newblock In {\em Proceedings of the 2019 Conference of the North {A}merican
  Chapter of the Association for Computational Linguistics: Human Language
  Technologies, Volume 1 (Long and Short Papers)}, pages 4149--4158,
  Minneapolis, Minnesota, June. Association for Computational Linguistics.

\bibitem[\protect\citename{{Vaswani} \bgroup et al.\egroup
  }2017]{NIPS2017_7181}
Ashish {Vaswani}, Noam {Shazeer}, Niki {Parmar}, Jakob {Uszkoreit}, Llion
  {Jones}, Aidan~N. {Gomez}, Lukasz {Kaiser}, and Illia {Polosukhin}.
\newblock 2017.
\newblock {Attention is All you Need}.
\newblock In I.~Guyon, U.~V. Luxburg, S.~Bengio, H.~Wallach, R.~Fergus,
  S.~Vishwanathan, and R.~Garnett, editors, {\em Advances in Neural Information
  Processing Systems 30}, pages 5998--6008. Curran Associates, Inc.

\bibitem[\protect\citename{Veli{\v{c}}kovi{\'{c}} \bgroup et al.\egroup
  }2018]{velickovic2018graph}
Petar Veli{\v{c}}kovi{\'{c}}, Guillem Cucurull, Arantxa Casanova, Adriana
  Romero, Pietro Li{\`{o}}, and Yoshua Bengio.
\newblock 2018.
\newblock {Graph Attention Networks}.
\newblock {\em International Conference on Learning Representations}.
\newblock accepted as poster.

\bibitem[\protect\citename{Wang \bgroup et al.\egroup
  }2019a]{wang-etal-2019-make}
Cunxiang Wang, Shuailong Liang, Yue Zhang, Xiaonan Li, and Tian Gao.
\newblock 2019a.
\newblock {Does it Make Sense? And Why? A Pilot Study for Sense Making and
  Explanation}.
\newblock In {\em Proceedings of the 57th Annual Meeting of the Association for
  Computational Linguistics}, pages 4020--4026, Florence, Italy, July.
  Association for Computational Linguistics.

\bibitem[\protect\citename{Wang \bgroup et al.\egroup
  }2019b]{DBLP:conf/aaai/WangKMYTACFMMW19}
Xiaoyan Wang, Pavan Kapanipathi, Ryan Musa, Mo~Yu, Kartik Talamadupula, Ibrahim
  Abdelaziz, Maria Chang, Achille Fokoue, Bassem Makni, Nicholas Mattei, and
  Michael Witbrock.
\newblock 2019b.
\newblock {Improving Natural Language Inference Using External Knowledge in the
  Science Questions Domain}.
\newblock In {\em The Thirty-Third {AAAI} Conference on Artificial
  Intelligence, {AAAI} 2019, The Thirty-First Innovative Applications of
  Artificial Intelligence Conference, {IAAI} 2019, The Ninth {AAAI} Symposium
  on Educational Advances in Artificial Intelligence, {EAAI} 2019, Honolulu,
  Hawaii, USA, January 27 - February 1, 2019}, pages 7208--7215. {AAAI} Press.

\bibitem[\protect\citename{Wang \bgroup et al.\egroup
  }2020]{wang-etal-2020-semeval}
Cunxiang Wang, Shuailong Liang, Yili Jin, Yilong Wang, Xiaodan Zhu, and Yue
  Zhang.
\newblock 2020.
\newblock {{S}em{E}val-2020 Task 4: Commonsense Validation and Explanation}.
\newblock In {\em Proceedings of The 14th International Workshop on Semantic
  Evaluation}. Association for Computational Linguistics.

\bibitem[\protect\citename{{Weissenborn} \bgroup et al.\egroup
  }2017]{Weissenborn2018DynamicIO}
Dirk {Weissenborn}, Tom{\'a}{\v{s}} {Ko{\v{c}}isk{\'y}}, and Chris {Dyer}.
\newblock 2017.
\newblock {Dynamic Integration of Background Knowledge in Neural NLU Systems}.
\newblock {\em arXiv e-prints}, page arXiv:1706.02596, June.

\bibitem[\protect\citename{Yang \bgroup et al.\egroup }2019]{NIPS2019_8812}
Zhilin Yang, Zihang Dai, Yiming Yang, Jaime Carbonell, Russ~R Salakhutdinov,
  and Quoc~V Le.
\newblock 2019.
\newblock {XLNet: Generalized Autoregressive Pretraining for Language
  Understanding}.
\newblock In {\em Advances in Neural Information Processing Systems 32}, pages
  5753--5763. Curran Associates, Inc.

\bibitem[\protect\citename{Zhong \bgroup et al.\egroup
  }2019]{10.1007/978-3-030-32233-5_2}
Wanjun Zhong, Duyu Tang, Nan Duan, Ming Zhou, Jiahai Wang, and Jian Yin.
\newblock 2019.
\newblock {Improving Question Answering by Commonsense-Based Pre-training}.
\newblock In Jie Tang, Min-Yen Kan, Dongyan Zhao, Sujian Li, and Hongying Zan,
  editors, {\em Natural Language Processing and Chinese Computing}, pages
  16--28, Cham. Springer International Publishing.

\end{thebibliography}

\nocite{2013arXiv1301.3781M} 
\nocite{kipf2017semi} 
\nocite{bosselut-etal-2019-comet} 
\nocite{DBLP:journals/corr/abs-1909-09743}  

\end{document}